\documentclass[conference]{IEEEtran}
\IEEEoverridecommandlockouts
\usepackage{cite}
\usepackage{amsmath,amssymb,amsfonts}
\usepackage{algorithmic}
\usepackage{graphicx}
\usepackage{textcomp}
\usepackage{xcolor}
\usepackage[table]{xcolor}
\usepackage{booktabs}
\usepackage{url}

\usepackage{subcaption}
\usepackage{multirow}
\def\BibTeX{{\rm B\kern-.05em{\sc i\kern-.025em b}\kern-.08em
    T\kern-.1667em\lower.7ex\hbox{E}\kern-.125emX}}
\begin{document}

\title{ICA-RAG: Information Completeness Guided Adaptive Retrieval-Augmented Generation for Disease Diagnosis\\
\thanks{* Equal contribution.}
\thanks{† Corresponding author.}
}


\author{
 \textbf{Jiawei He\textsuperscript{1*}},
 \textbf{Mingyi Jia\textsuperscript{1*}},
 \textbf{Zhihao Jia\textsuperscript{1}},
 \textbf{Junwen Duan\textsuperscript{1†}}
 \textbf{Yan Song\textsuperscript{2}},
 \textbf{Jianxin Wang\textsuperscript{1}},
\\
 \textsuperscript{1}School of Computer Science and
Engineering, Central South University \\
 \textsuperscript{2}University of Science and Technology of China,
\\
\texttt{\{hejiawei, jiamingyi, zhihaojia, jwduan\}@csu.edu.cn}
\\
\texttt{\ clksong@gmial.com, jxwang@mail.csu.edu.cn}
}

\maketitle

\begin{abstract}
Retrieval-Augmented Large Language Models (LLMs), which integrate external knowledge, have shown remarkable performance in medical domains, including clinical diagnosis. However, existing RAG methods often struggle to tailor retrieval strategies to diagnostic difficulty and input sample informativeness. This limitation leads to excessive and often unnecessary retrieval, impairing computational efficiency and increasing the risk of introducing noise that can degrade diagnostic accuracy. To address this, we propose ICA-RAG (\textbf{I}nformation \textbf{C}ompleteness Guided \textbf{A}daptive \textbf{R}etrieval-\textbf{A}ugmented \textbf{G}eneration), a novel framework for enhancing RAG reliability in disease diagnosis. ICA-RAG utilizes an adaptive control module to assess the necessity of retrieval based on the input's information completeness. By optimizing retrieval and incorporating knowledge filtering, ICA-RAG better aligns retrieval operations with clinical requirements. Experiments on three Chinese electronic medical record datasets demonstrate that ICA-RAG significantly outperforms baseline methods, highlighting its effectiveness in clinical diagnosis.
\end{abstract}


\section{Introduction}
Large Language Models (LLMs) \cite{openai2023gpt4,llmdd160_saab2024capabilities} have demonstrated exceptional capabilities in medical tasks, including clinical diagnosis~\cite{zhou2024llmddoverview}. However, their adoption faces challenges such as hallucination—the generation of plausible but incorrect information~\cite{maynez2020faithfulness, llmdd245_huang2023survey}—and the resource-intensive nature of knowledge updates~\cite{zhang2023large, kasai2024realtime}. 
Retrieval-augmented generation (RAG)~\cite{lewis2020retrieval} offers a solution by integrating trustworthy external documents to reduce hallucinations and ensure up-to-date information.

\begin{figure}[t]
	\centering
	\includegraphics[width=\columnwidth]{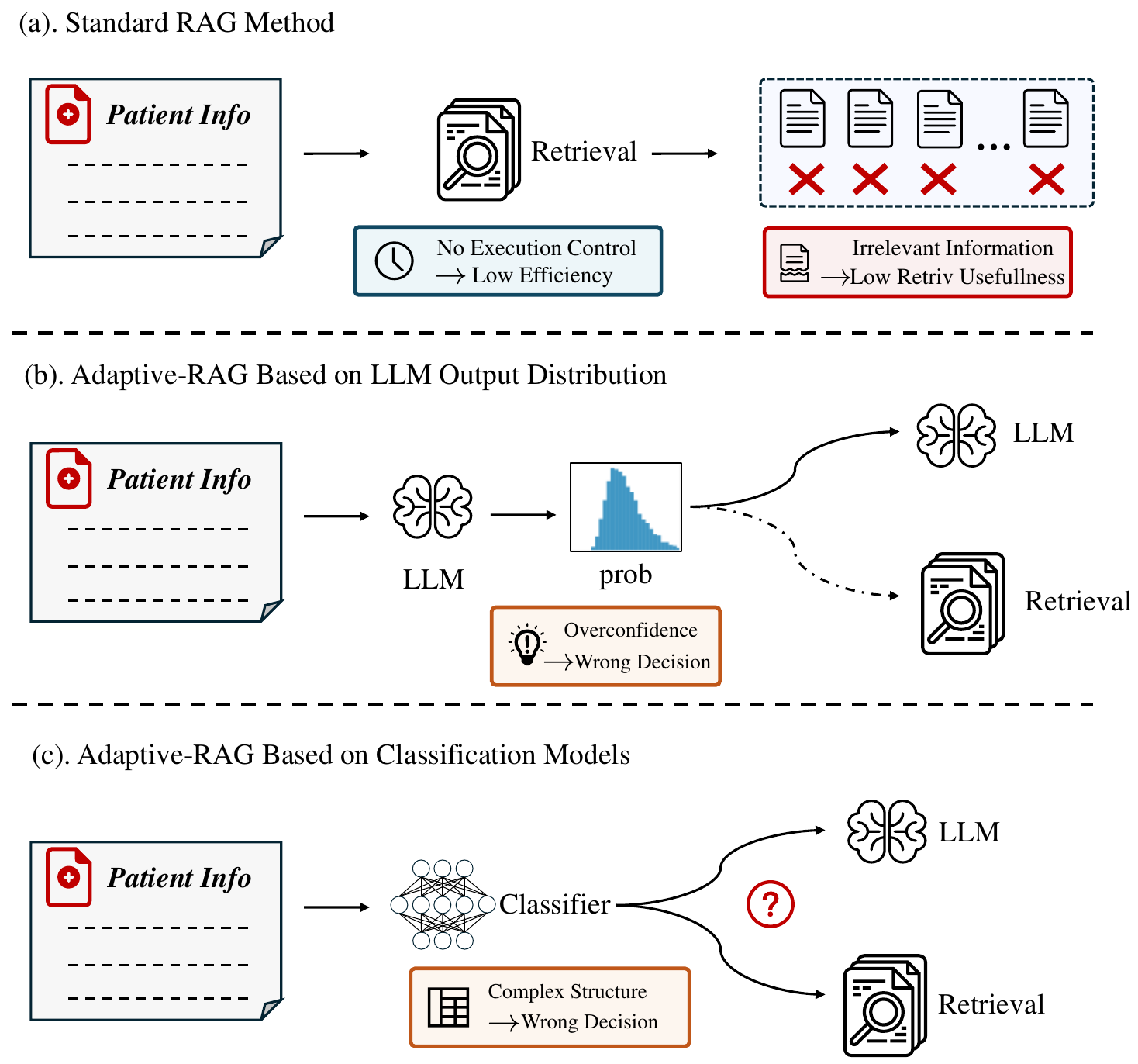}
	\caption{Illustration of three different RAG paradigms for solving clinical diagnosis task.}
	\label{example1}
\end{figure}

While researchers have widely explored RAG to enhance LLM accuracy in high-risk domains~\cite{llmdd_ovewview_zhou2024large}, not all medical cases require this approach. Many common diseases or cases with mild symptoms and clear diagnoses can be accurately addressed without retrieval~\cite{jeong2024adaptive}. However, most existing RAG methods lack selective retrieval logic, instead performing retrievals for all queries indiscriminately. This approach not only increases computational and time costs but may also introduce errors through low-quality retrievals (as shown in Fig.~\ref{example1}-(a)), potentially degrading rather than improving performance.

To improve the efficiency of retrieval systems, researchers have proposed adaptive RAG paradigms~\cite{jeong2024adaptive, su2024dragin, yao2024seakr}, which establish control logic to activate the retrieval system only when certain conditions are met. There are two common approaches in these paradigms: (1) setting judgment conditions based on LLM's output or probability distributions~\cite{yao2024seakr, su2024dragin}; (2) training a relatively smaller judgment model to determine whether to perform retrieval at a lower cost~\cite{jeong2024adaptive}. Fig.~\ref{example1}-(b) and (c) provide corresponding examples of these approaches.

However, the former approach has limitations as LLMs tend to be overconfident, generating high-confidence probability distributions even when lacking relevant knowledge~\cite{huang2023large, xu2024perils}. Additionally, these methods typically require access to LLM output probability distributions (logits), limiting adaptability for API services or closed-source model applications. While the latter approach relies heavily on input content characteristics. For instance, Jeong et al.~\cite{jeong2024adaptive} define "simple questions" as single-hop queries (e.g., "When is Michael F. Phelps's birthday?") and "difficult questions" as multi-hop queries (e.g., "What currency is used in Bill Gates's birthplace?"). Such question-answering tasks have distinct difficulty gradients, making them relatively easy for models to differentiate.

Unlike single-hop or multi-hop question answering tasks, input texts in the medical domain typically do not exhibit obvious structural patterns that can be captured, making it extremely challenging for smaller language models to understand the difficulty of answering them. Therefore, the successful experiences from this approach cannot be directly transferred to other tasks.

Based on above analysis, we proposed a disease diagnosis approach ICA-RAG (\textbf{I}nformation \textbf{C}ompleteness Guided \textbf{A}daptive \textbf{R}etrieval-\textbf{A}ugmented \textbf{G}eneration), using adaptive retrieval decision optimization, specifically tailored for complex structured and long-context medical texts. The core innovation introduces a retrieval decision optimization module based on input information completeness. This module segments long inputs into text units, employs a classification model to predict each unit's importance, and calculates global information completeness to determine retrieval necessity. Since the classifier already identifies important text units, these can be prioritized during retrieval, minimizing interference from irrelevant information. Through a single prediction round, this module achieves both retrieval decision optimization and query selection, effectively addressing the limitations in existing RAG paradigms.

Our main contributions are as follows:
\begin{itemize}
    \item We propose ICA-RAG, a framework for adaptive retrieval-augmented disease diagnosis without the need for tuning backbone LLMs.
    \item We desgined a novel data annotation methodology that employs masking operations to elicit varied responses from LLMs, thereby acquiring label information. Concurrently, we have optimized the retrieval process to better accommodate clinical scenarios with complex context.
    \item We conducted extensive experiments on three Chinese EMR datasets to demonstrate the effectiveness of our ICA-RAG framework.
\end{itemize}

\section{Related Work}

\subsection{RAG in Clinical Disease Diagnosis}
To improve diagnostic accuracy, model reliability, and reduce hallucination issues without retraining, recent studies widely adopt Retrieval-Augmented Generation (RAG) to integrate external medical knowledge \cite{llmdd69_wen2023mindmap,llmdd70_wu2024guiding,llmdd79_shi2023retrieval,llmdd141_thompson2023large,llmdd142_zhao2024heterogeneous}. Most research uses basic retrieval methods \cite{llmdd74_ge2024development,llmdd79_shi2023retrieval,llmdd140_zhang2023integrating,llmdd142_zhao2024heterogeneous,llmdd148_oniani2024enhancing}, typically leveraging embedding models to encode external knowledge and task queries into vector representations. Relevant knowledge is retrieved via vector similarity and used in LLMs through tailored prompts for diagnosis generation. Besides, knowledge graphs are also widely employed \cite{llmdd69_wen2023mindmap,llmdd70_wu2024guiding,llmdd71_gao2023large}.

\begin{figure*}[t]
	\centering
	\includegraphics[width=\textwidth]{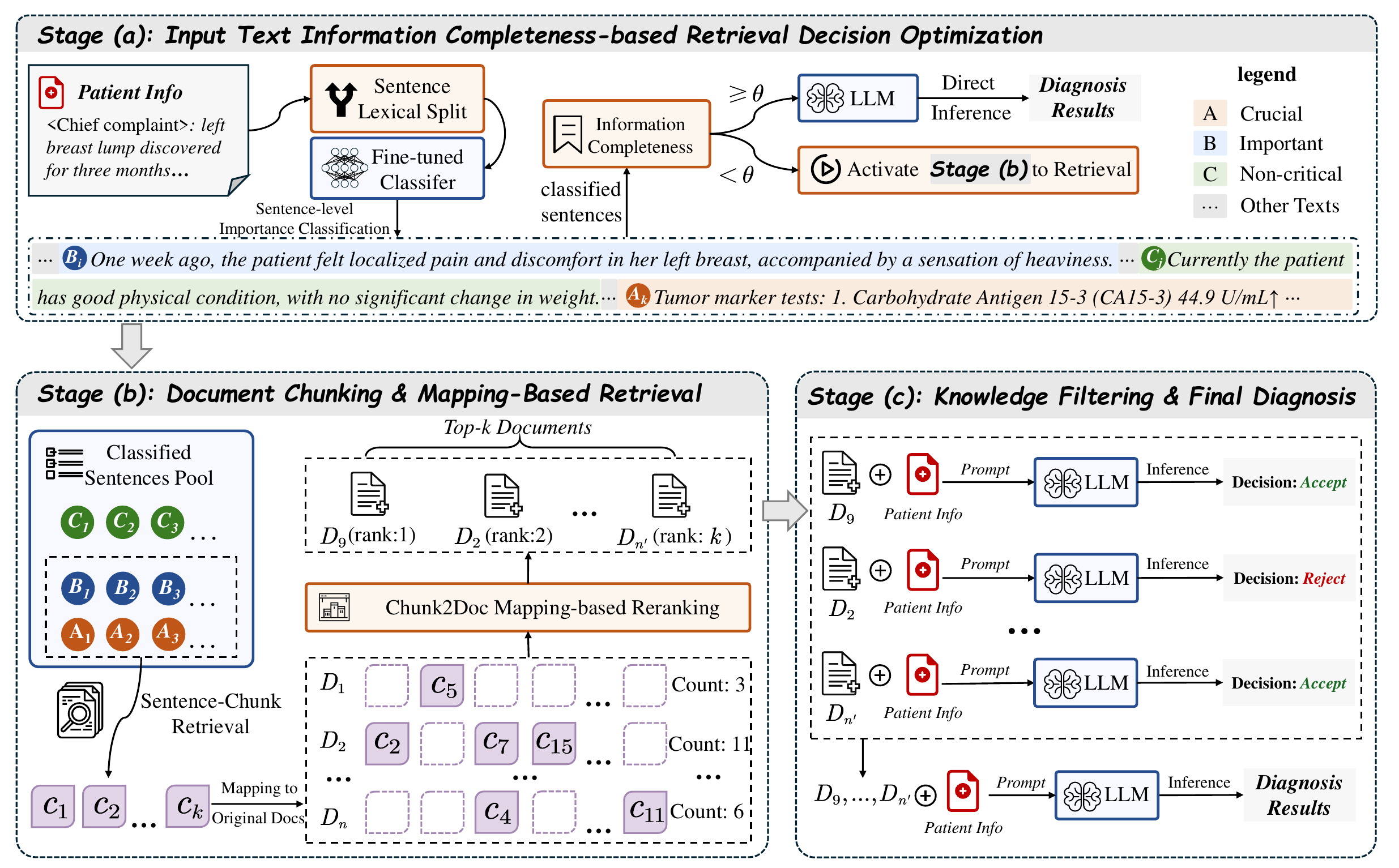}
	\caption{The overall architecture of our proposed framework ICA-RAG. It consists of three stages. Stage(a) involves inference \& Retrieval Decision Making Based on Fine-Grained Information Density. Stage (b) focuses on knowledge retrieval and integration. Note that Stage (b) and (c) is activated only when the score computed in Stage (a) falls below a predefined threshold.}
	\label{workflow}
\end{figure*}

\subsection{Adaptive-RAG}
Adaptive Retrieval-Augmented Generation (RAG) dynamically determines whether a large language model (LLM) requires external knowledge retrieval to mitigate inaccuracies. FLARE~\cite{jiang2023active} and DRAGIN~\cite{su2024dragin} activate search engines when the LLM generates low-confidence tokens. Wang et al.~\cite{wang2024llms} use a prompting mechanism for LLMs to autonomously decide on retrieval. Self-Awareness-Guided Generation~\cite{wang2023self} trains a classifier to assess output authenticity, while Adaptive-RAG~\cite{jeong2024adaptive} evaluates query complexity to determine retrieval necessity. Mallen et al.~\cite{mallen2023not} propose activating retrieval based on entity frequency in queries, though this may fail for complex, multi-step reasoning tasks. Asai et al. introduce Self-RAG~\cite{asai2023self}, which trains a model to dynamically retrieve, critique, and generate text.

\begin{figure*}[t]
	\centering
	\includegraphics[width=\textwidth]{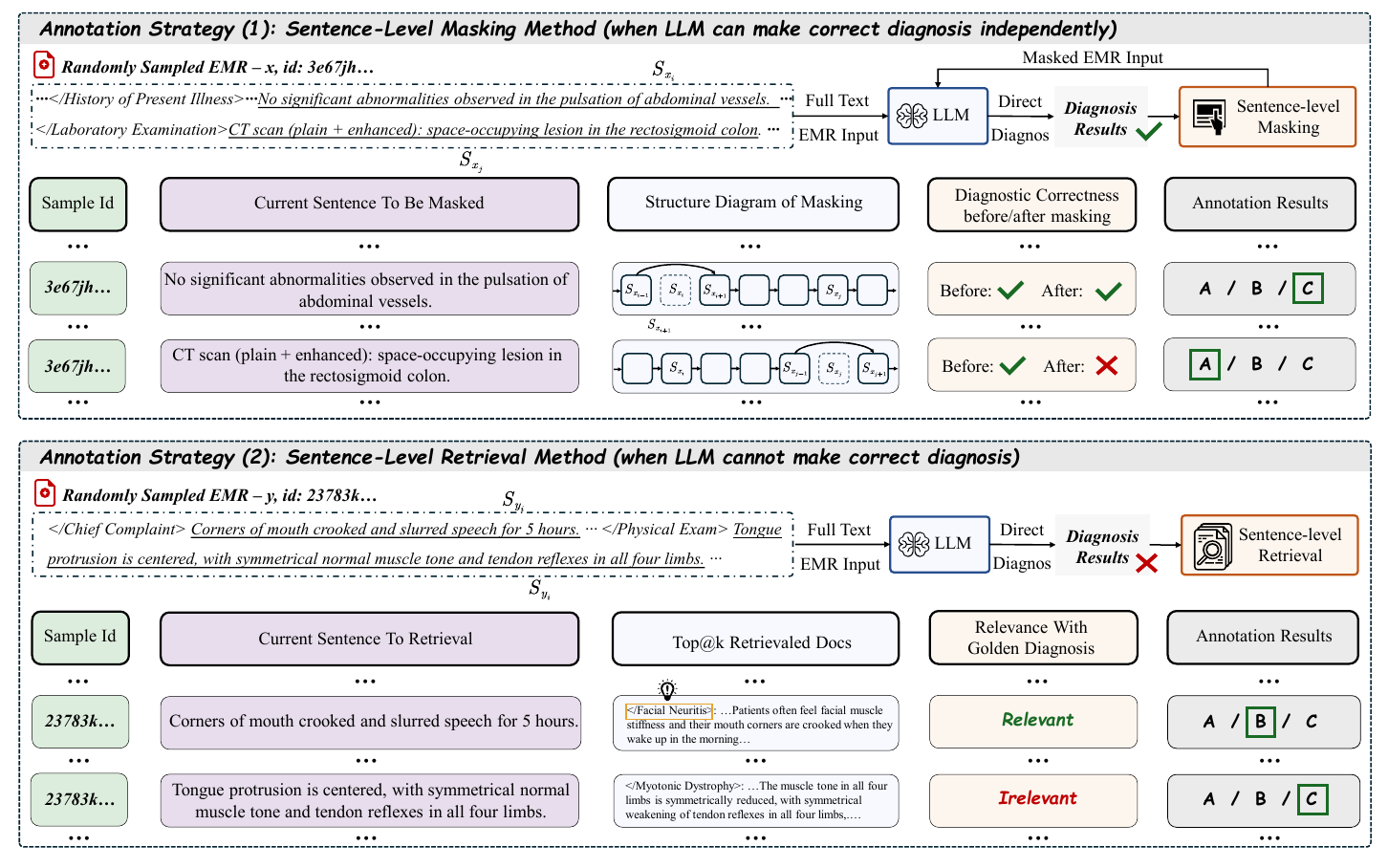}
	\caption{Details of our proposed annotation strategy. During the annotation process, we adopt different annotation strategies based on the responses generated by the LLM.}
	\label{annotate}
\end{figure*}

\section{Methods}
In this section, we first present the formal definition of disease diagnosis task and the task settings for adaptive-RAG-based disease diagnosis. Then we will introduce the details of each components of our proposed ICA-RAG framework.

\subsection{Preliminaries}
\noindent \textbf{Direct Disease Diagnosis via LLM}: Given a token sequence $\mathbf{x} = [x_1, x_2, \ldots, x_n]$ representing input text, LLM-based text generation can be formalized as $\mathbf{y} = \mathrm{LLM}(\mathbf{x}, prompt)$, where $prompt$ is a task-specific template and $\mathbf{y} = [y_1, y_2, \ldots, y_n]$ is the generated output. For disease diagnosis, the input $\mathbf{x}$ is patient information $\mathcal{Q}$, and the output $\mathbf{y}$ is the predicted diagnosis $\hat{\mathcal{D}}$, formalized as: $\hat{\mathcal{D}} = \mathrm{LLM}(\mathcal{Q}, prompt)$.

\vspace{0.4\baselineskip}

\noindent\textbf{RAG-based Disease Diagnosis}: This approach retrieves relevant knowledge ${d}$ from an external knowledge source $\mathcal{K}$ using a retrieval module $\mathsf{Retriever}$. The diagnosis is then generated by incorporating this knowledge: $\hat{\mathcal{D}} = \mathrm{LLM}(\mathcal{Q},d, prompt)$, where $d = \mathsf{Retriever}(\mathcal{K},\mathcal{Q})$. In this paper, we use a document knowledge base $\mathsf{KB}$ as the external knowledge source.

\vspace{0.4\baselineskip}

\noindent\textbf{Adaptive-RAG-based Disease Diagnosis}: This paradigm introduces a control function $F$ that evaluates input $\mathcal{Q}$ to determine whether retrieval is necessary:
\begin{equation}
	\hat{\mathcal{D}} = 
	\begin{cases} 
	\mathrm{LLM}(\mathcal{Q}, prompt), & \text{if } F(\mathcal{Q})=\langle Activate \rangle \\
	\mathrm{LLM}(\mathcal{Q}, d, prompt), & \text{otherwise}
	\end{cases}
\end{equation}
where $d = \mathsf{Retriever}(\mathcal{K},\mathcal{Q})$. The control function $F$ can be implemented through various approaches, such as LLM token probability distributions, confidence levels, or a smaller trained decision model.

\subsection{Retrieval Decision Optimization Based on Input Information Completeness} \label{adaptive_control}

\vspace{0.5ex}
\textit{1) Calculation of Input Information Completeness}
\vspace{0.5ex}

Although smaller language models can evaluate the complexity of input questions and make retrieval decisions~\cite{jeong2024adaptive}, they struggle with long, complex medical diagnostic contexts. These models often rely on superficial features rather than semantic understanding when processing extensive inputs. Training larger models specifically for this purpose~\cite{asai2023self} is resource-intensive and contradicts RAG paradigm objectives.

To address this limitation, we segments the input $\mathcal{Q}$ into manageable text units (defaulting to sentences): $\mathcal{Q} = \{s_i\}_{i=1}^n$, and trains a language model $\mathsf{Classifier}$ to predict each unit's importance. As shown in the left half of Fig.~\ref{workflow}.Stage~(a):
\begin{equation}
	l_i = \mathsf{Classifier}(s_i) \quad \forall i \in \{1, 2, ..., n\}
\end{equation}
This approach transforms complex document comprehension into simpler sentence-level tasks. Each text unit $s_i$ receives one of three labels $\{\mathsf{A, B, C}\}$: $\mathsf{A}$ for information critical to diagnostic decisions, $\mathsf{B}$ for information that positively contributes to retrieval without directly inferring the correct result, and $\mathsf{C}$ for relatively unimportant information.

Based on the classification results, we calculate the global information completeness of input $\mathcal{Q}$ as follows:
\begin{align}
	I_{\text{norm}}(\mathcal{Q}) & = \frac{1}{\alpha \cdot n} \sum_{i=1}^n \Big( \alpha \cdot \mathbb{I}(l_i = \mathsf{A}) \\ 
	& + \beta \cdot \mathbb{I}(l_i = \mathsf{B}) + \gamma \cdot \mathbb{I}(l_i = \mathsf{C}) \Big)
\end{align}
where $l_i$ is the classification result of text unit $s_i$, $\alpha$, $\beta$, and $\gamma$ are weights for the three category labels, and $\mathbb{I}(\cdot)$ is an indicator function that returns 1 when the condition is true and 0 otherwise. The denominator $\alpha \cdot n$ in the equation represents the maximum information completeness (when all sentences are classified as $\mathsf{A}$), serving as normalization. Inputs with more critical clues increase the LLM's potential for accurate diagnosis. When $I_{\text{norm}}$ exceeds $\theta_1$, the input contains sufficient information for direct diagnosis:
\begin{equation}
	\mathcal{D}_{final} = \mathrm{LLM}(\mathcal{Q}, prompt_{diag})
\end{equation}

If $I_{\text{norm}}$ falls between $\theta_1$ and $\theta_2$, the retrieval program activates (see Section~\ref{retrieval_process}). When $I_{\text{norm}}$ is below $\theta_2$, a warning signal is issued alongside normal retrieval and reasoning, indicating sparse critical information and potential misdiagnosis risk. 

\vspace{0.5ex}
\textit{2) Annotation Method for Classifier Training Data Based on Masking Strategy}
\vspace{0.5ex}

In the first part of this subsection, we detailed the implementation approach of the retrieval decision optimization module based on input information completeness. However, due to the lack of annotated datasets meeting our requirements for training importance classification models, we propose a simple yet effective strategy to construct and annotate training datasets. Inspired by dynamic token deletion from single-stage Weakly Supervised Rationale Extraction~\cite{jiang2023you}, we annotate the importance category of each text unit by sequentially masking them, as illustrated in Fig.~\ref{annotate}. 

For a given input $\mathcal{Q}$, we set the 
doctor's 
diagnostic result $\bar{\mathcal{R}}$ as the reference answer, then segment $\mathcal{Q}$ into multiple text units $\mathcal{Q} = [s_1, s_2, \ldots, s_n]$. We sequentially mask each text unit $s_i$ to obtain the masked input $\mathcal{Q'} = [s_1, s_2, \ldots, s_{i-1}, s_{i+1}, \ldots, s_n]$. The LLM then performs diagnostic reasoning based on both $\mathcal{Q}$ and $\mathcal{Q'}$ to generate predicted diagnoses:
\begin{align}
	\hat{\mathcal{D}} &= \mathrm{LLM}(\mathcal{Q}, {prompt}_{diag}) \label{eq:first} \\
	\hat{\mathcal{D'}} &= \mathrm{LLM}(\mathcal{Q'}, {prompt}_{diag}) \label{eq:second}
\end{align}
where $\hat{\mathcal{D}}$ and $\hat{\mathcal{D'}}$ represent the diagnostic results based on the complete and masked inputs, respectively. By comparing these diagnostic results with the standard answer $\bar{\mathcal{R}}$, we present two annotation strategies:

\textbf{Annotation Strategy (1).} If $\hat{\mathcal{D}} \approx \bar{\mathcal{R}}$ (the LLM makes correct predictions with complete input): If $\hat{\mathcal{D'}} \approx \bar{\mathcal{R}}$ also holds, indicating that masking $s_i$ does not significantly impact the reasoning process, then $s_i$ is labeled as $\mathsf{C}$ (non-critical information). If $\hat{\mathcal{D'}}$ differs from $\hat{\mathcal{D}}$ resulting in an incorrect diagnosis, $s_i$ is labeled as $\mathsf{A}$ (critical diagnostic information). This strategy is illustrated in the upper part of Fig.~\ref{annotate}.

\textbf{Annotation Strategy (2).} If $\hat{\mathcal{D}} \neq \bar{\mathcal{R}}$ (the LLM cannot make correct predictions with complete input): In this case, we implement annotation by searching the knowledge base. We use \(s_i\) as the retrieval query with the BM25 method. If documents corresponding to the disease in $\bar{\mathcal{R}}$ can be retrieved, \(s_i\) is labeled as $\mathsf{B}$ (valuable diagnostic information). Otherwise, \(s_i\) is labeled as $\mathsf{C}$ (low importance). This strategy is illustrated in the lower part of Fig.~\ref{annotate}.

\subsection{Knowledge Retrieval and Reranking Based on Document Segmentation and Mapping}
\label{retrieval_process}

Considering the complex structures, large context spans, and semantic discontinuities in clinical texts, we adapt the RAG process following Zhao et al.~\cite{zhao2024longrag}. This approach divides documents in the knowledge base $\mathsf{KB}$ into text chunks with length restrictions, using sentences as the minimum segmentation unit. Fig.~\ref{workflow}.Stage-b illustrates our retrieval and reranking workflow.

Given an input text $\mathcal{Q} = \{s_i\}_{i=1}^n$ with $n$ sentences, we first perform sentence-level importance classification and calculate overall information completeness $I_{\text{norm}}$ as described in Section~\ref{adaptive_control} (1). When $I_{\text{norm}}$ falls below a preset threshold, the retrieval module Stage-b activates. To optimize retrieval efficiency, we only retain sentences with $\mathsf{label}=\mathsf{A}$ and $\mathsf{label}=\mathsf{B}$, excluding those with $\mathsf{label}=\mathsf{C}$ (shown on the left side of Fig.~\ref{workflow}.Stage-b). This exclusion is justified as $\mathsf{label}=\mathsf{C}$ sentences typically contain non-pathological descriptions that contribute minimally to retrieval and may introduce noise.

The retrieval algorithm operates on knowledge base $\mathsf{KB}$ through chunk-level retrieval and document-level reranking. Each sentence $s_i \in \mathcal{Q}$ serves as a query to retrieve the top $m$ relevant text chunks:
\begin{equation}
	\mathcal{C}_i = \mathsf{Retriever}(s_i, m) \quad \forall i \in \{1, 2, \dots, n\}
\end{equation}
where $\mathcal{C}_i = \{c_{i,j}\}_{j=1}^m$, and $c_{i,j}$ is the $j$-th chunk retrieved using $s_i$. All text chunk sets are merged into $\mathcal{C} = \bigcup_{i=1}^n \mathcal{C}_i$. Each chunk $c \in \mathcal{C}$ is mapped to its original document $doc \in \mathsf{KB}$. For each document ${doc}$, a score $S_{{doc}}$ counts the number of retrieved chunks from that document:
\begin{equation}
	S_{doc} = \sum_{c \in \mathcal{C}} \mathbb{I}(c \in {doc})
\end{equation}
where $\mathbb{I}(\cdot)$ is the indicator function, returning 1 if $c$ belongs to $doc$ and 0 otherwise. Documents are reranked based on $S_{doc}$, and the top $k$ documents with highest scores are selected as the final retrieval results: $\mathcal{K}_{rerank} = \{{doc}_{(1)}, {doc}_{(2)}, \dots, {doc}_{(k)}\}$, where ${doc}_{(l)}$ represents the document with the $l$-th highest score.

\subsection{Knowledge Filtering and Diagnosis Generation Based on Prompt Guidance}

Despite optimizing the retrieval process, retrieved documents may not always be relevant, particularly in clinical diagnostic tasks requiring complex reasoning. Drawing inspiration from medical "differential diagnosis" procedures, where doctors examine potentially confusing diseases based on patient symptoms and test results, we designed a prompt template \(prompt_\textit{diff}\) to filter irrelevant information. This template guides the LLM to identify conflicts between patient information and document descriptions, determining which documents to retain. The process is illustrated in Fig.~\ref{workflow}.Stage-c.

Given the reranked knowledge document set $\mathcal{K}_{rerank}=\{{doc}_{(1)}, {doc}_{(2)}, \dots, {doc}_{(k)}\}$, we filter documents by evaluating their relevance to the diagnosis. The filtering function $V(\mathcal{Q}, doc_{(i)}, prompt_\textit{diff})$ is defined as:
\begin{equation}
	V(\mathcal{Q}, doc_{(i)}, prompt_\textit{diff}) = 
	\begin{cases} 
	\text{True}, &\text{if } \langle\text{support}\rangle \\
	\text{False}, &\text{otherwise}
	\end{cases}
\end{equation}
where $\langle\text{support}\rangle$ represents the LLM output when provided with query $\mathcal{Q}$, document $doc_{(i)}$, and prompt template $prompt_\textit{diff}$. The term $\langle\text{support}\rangle$ indicates that the LLM determines $doc_{(i)}$ is critical for diagnosis. The final reference knowledge document set $\mathcal{K}^*$ retains only documents that satisfy the filtering condition:
\begin{equation}
	\mathcal{K}^* = \{doc_{(i)} \in \mathcal{K}_{rerank} \mid V(\cdot, doc_{(i)}, \cdot) = \text{True}\}
\end{equation}

The final RAG-based diagnostic process is formalized as:
\begin{equation}
	\mathcal{D}_{final} = \mathrm{LLM}(\mathcal{Q}, \mathcal{K}^*, prompt_{rag})
\end{equation}
where $prompt_{rag}$ represents the RAG-based diagnostic prompt template.

\section{Experiments}
\subsection{Datasets}
We evaluated our framework using three Chinese EMR datasets: CMEMR~\cite{mediKAL}, ClinicalBench~\cite{yan2024clinicalbench}, and CMB-Clin~\cite{cmb}, to assess its ability in analyzing complex clinical information and making accurate diagnoses. For the task setup, all three datasets are configured into end-to-end diagnostic tasks, where patient information (such as chief complaints, medical history, and examination findings) serves as input, with physicians' diagnostic conclusions as ground-truth labels.

\subsection{Experimental Settings}
\noindent \textbf{Baseline Methods.} We compare our approach with three categories of methods. 
(1) Non-Retrieval methods: We include Chain-of-Thought (CoT)~\cite{wei2022chain}, Self-Consistent Chain of Thought(Sc-CoT)~\cite{llmdd113_wang2022self} and Atypical Prompting~\cite{Atypical}.
(2) Standard-Retrieval methods: We include two representative RAG methods: $\text{RAG}^\text{2}$~(Rationale-Guided RAG)\cite{sohn2024rationale} and LongRAG~\cite{zhao2024longrag}.
(3) Adaptive-Retrieval methods: We include Adaptive-RAG~\cite{jeong2024adaptive}, DRAGIN~\cite{su2024dragin}, and SEAKR~\cite{yao2024seakr}.

\vspace{0.5ex}
\noindent \textbf{Evaluation Metric.} Following \cite{fan2024ai-hospital}, we use the International Classification of Diseases (ICD-10) \cite{percy1990ICD} to standardize disease terminologies. We extract disease entities from diagnostic results and EMR labels, then perform fuzzy matching with a threshold of 0.5 to link them to ICD-10, creating normalized sets $S_{\mathcal{\hat{D}}}$ and $S_{\mathcal{R}}$. These sets are used to calculate set-level metrics Precision, Recall, and F1-score.

\vspace{0.5ex}
\noindent \textbf{Implementation Details.} We choose qwen2.5-7B-instruct as the backbone model for inference in our experiments by default. For the classifier we choose BERT-base-Chinese~\cite{llmsv3_devlin2018bert}. For the retriever we use BM25\cite{bm25} by default. For the external knowledge corpus we use CMKD~(Clinical Medicine Knowledge Database)\footnote{\url{http://cmkd.juhe.com.cn/}}.

\subsection{Main Results}
Our experiments evaluate the framework against baselines on three Chinese EMR datasets. Table~\ref{main_exp} highlights key findings:

(1) ICA-RAG demonstrates consistent performance across all benchmark datasets, achieving optimal or near-optimal F1 scores compared to baseline methods.

(2) Compared to LongRAG, a superior conventional retrieval approach, ICA-RAG improves Set-level F1 values by 1.81\%, 1.54\%, and 1.72\% respectively on the three datasets. This indicates that standard RAG methods without retrieval decision optimization rely excessively on knowledge base quality in complex disease diagnosis scenarios. They initiate retrieval even when LLMs can independently complete diagnoses, reducing efficiency and potentially introducing errors. 

\begin{table*}[t]
\caption{Experimental results on CMEMR, ClinicalBench and CMB-Clin datasets. Bold indicates the best performances and the second-best performances are underlined.}
\centering
\begin{tabular}{llllllllll}
\toprule
\hline
\multicolumn{1}{c}{\multirow{2}{*}{Method}} & \multicolumn{3}{c}{CMEMR} & \multicolumn{3}{c}{ClinicalBench} & \multicolumn{3}{c}{CMB-Clin} \\ \cmidrule(r){2-4} \cmidrule(r){5-7} \cmidrule(r){8-10}
\multicolumn{1}{c}{} & R~(\%)& P~(\%) & F1~(\%) & R~(\%) & P~(\%) & F1~(\%) & R~(\%) & P~(\%) & F1~(\%) \\ \midrule
\multicolumn{10}{l}{\textbf{\textit{Non-Retrieval Methods}}} \\
		CoT& 49.09 & \underline{48.56} & {48.82} & 44.12 & 34.09 & 38.46 & 68.03 & 42.27 & 52.14 \\
		SC-CoT & 49.49 & 48.21 & 48.84 & 42.74 & 33.41 & 37.50 & {69.27} & 43.43 & 53.39 \\
		ATP & {49.68} & 47.72 & 48.68 & 43.01 & 33.82 & 37.87 & \underline{70.83} & \underline{44.73} & \textbf{54.83} \\ \midrule
		\multicolumn{10}{l}{\textbf{\textit{Standard Retrieval Methods}}} \\ 
		$\text{RAG}^\text{2}$ & 47.13 & 44.34 & 45.69 & 42.43 & 34.57 & 38.10 & 58.33 &35.29  & 43.98 \\
		LongRAG & {49.85} & 48.31 & 49.07 & \underline{44.65} & 35.02 & \underline{39.25} & 69.44 & 41.32 & 51.81 \\
		\midrule
		\multicolumn{10}{l}{\textbf{\textit{Adaptive Retrieval Methods}}} \\
        \renewcommand{\arraystretch}{1.1}
		DRAGIN & 47.09 & 46.92 &  47.00& 43.67 & \underline{35.54} & {39.19} & 59.72 & 36.13 & 45.03 \\ 
		Adaptive-RAG & \underline{50.23} & 48.35 & \underline{49.27} & 42.23 & 34.61 & 38.04 & 65.37 & \textbf{45.20} & 53.44 \\ 
		SEAKR  & 47.37 & 45.90 & 46.62 & 40.66 & 33.13 & 36.51 & 59.60 & 34.34 & 43.57 \\
		\hline
        \addlinespace[2pt]	
		ICA-RAG~(ours) & \textbf{53.42} & \textbf{48.58} & \textbf{50.88}& \textbf{46.63} & \textbf{36.24} & \textbf{40.79} & \textbf{71.62} & 42.74 & \underline{53.53} \\ 
 \bottomrule
\end{tabular}
\label{main_exp}
\end{table*}

(3) ICA-RAG outperforms other adaptive RAG methods significantly. Compared to the best-performing Adaptive-RAG method, ICA-RAG exhibits enhanced robustness when handling structurally complex and long context inputs due to its adaptive decision-making based on local-to-global information completeness calculations. Most other baselines, on the other hand, are  designed primarily for simpler question answering tasks, so their performance fluctuations when applied to disease diagnosis without appropriate adaptations.

\subsection{Ablation Study}

\begin{table}[b]
	\centering
	\caption{Ablation study on CMEMR dataset. \emph{w/o} denotes  removing the corresponding module.}
    \renewcommand{\arraystretch}{1.1}
    \begin{tabular}{>{\centering\arraybackslash}p{2.5cm}lccc}
		\toprule
		Method & R~(\%) & P~(\%) & F1~(\%) \\ \hline
		ICA-RAG & \textbf{53.42} & \textbf{48.58} & \textbf{50.88}  \\
		\emph{w/o} Decision & 49.74 & 46.52 & 48.07 \\
		\emph{w/o} Chunk & 52.26 & 47.53 & 49.78 \\
		\emph{w/o} M-rerank & 52.22 & 47.20 & 49.59 \\
		\emph{w/o} Diff & 52.70 & 47.29 & 49.85 \\
		\bottomrule
	\end{tabular}
	\label{idea2_ablation}
\end{table}

To analyze the contribution of different modules in ICA-RAG to its performance, we conducted ablation experiments on the CMEMR dataset: (a) \emph{w/o} Decision: removing the retrieval decision optimization module; (b) \emph{w/o} Chunk: replacing ICA-RAG's document segmentation and mapping-based knowledge retrieval with direct retrieval of complete documents; (c) \emph{w/o} M-rerank (Mapping-based Rerank): replacing ICA-RAG's text chunk mapping-based reranking with the bge-reranker-v2-m3 model; (d) \emph{w/o} Diff: removing the LLM knowledge filtering module based on differential diagnosis prompting. The results are shown in Table~\ref{idea2_ablation}, leading to the following conclusions:

(1) Without the retrieval decision optimization module, ICA-RAG's F1 value dropped by 2.81\%. This occurs because all inputs undergo retrieval indiscriminately, forcing samples that LLM could diagnose independently to undergo unnecessary retrieval, reducing efficiency and increasing error risk from irrelevant information.

(2) Replacing ICA-RAG's document segmentation and mapping-based retrieval with original retrieval decreased F1 by 1.1\%. This demonstrates that general RAG methods struggle with sparse information distribution and semantic incoherence in clinical texts, hampering accurate matching between inputs and knowledge base documents.

(3) Substituting ICA-RAG's reranking method with the bge-reranker-v2-m3 model reduced performance, validating ICA-RAG's reranking design. ICA-RAG's approach relies solely on numerical calculations from retrieval results without additional models, reducing memory overhead while maintaining higher compatibility with the retrieval workflow.

(4) Removing the differential diagnosis-based knowledge filtering mechanism meant all retrieved documents were provided to the LLM without discrimination. This increased the difficulty of LLM's reasoning and raised the probability of exceeding input length limits, negatively impacting overall performance.

	
	

\begin{figure}[t]
    \centering
    
    \subfloat[Average Time Cost on CMEMR dataset.\label{fig:time}]{
        \includegraphics[width=0.85\columnwidth]{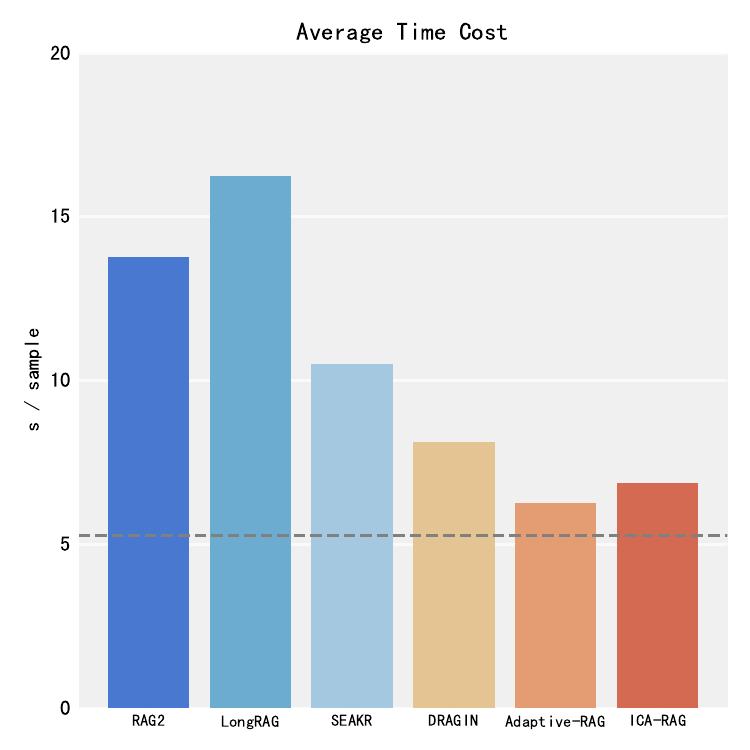}
    }\\
    \vspace{0.8em} 
    
    \subfloat[Diagnostic Performance on CMEMR dataset.\label{fig:performance}]{
        \includegraphics[width=0.85\columnwidth]{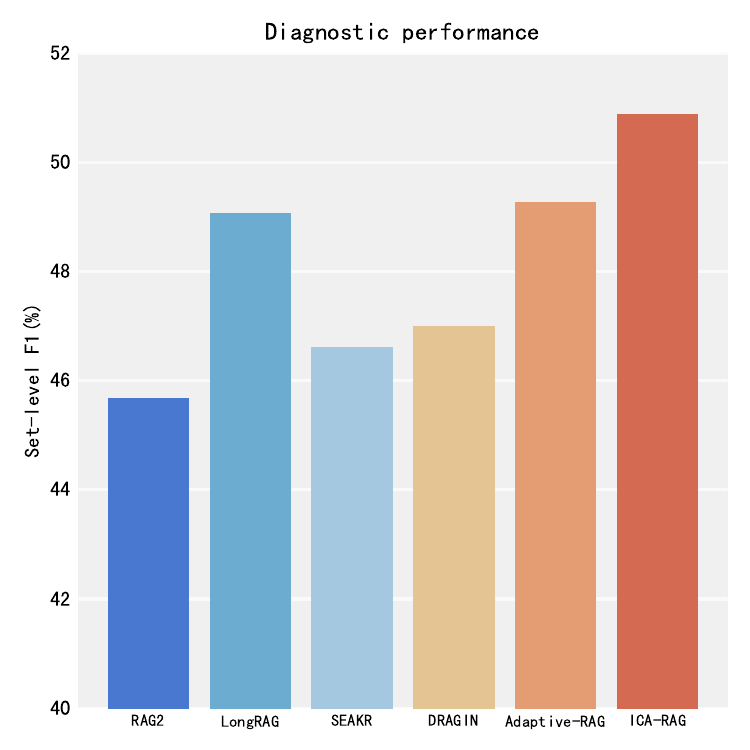}
    }
    
    \caption{A Comparative Analysis of Computational Time Expenditure and Diagnostic Performance Between the Proposed Method and Selected Baseline Methods on the CMEMR Dataset.}
    \label{time_and_performance}
\end{figure}

\subsection{Analysis of Retrieval Decision Optimization Effects}
We compare our method with other retrieval-based baselines in terms of efficiency and diagnostic performance, as shown in Fig.~\ref{time_and_performance}. Based on the experimental results, the following conclusions can be drawn:

(1) As illustrated in Fig.~\ref{time_and_performance}.a, our method demonstrates significant time efficiency advantages compared to non-adaptive RAG methods ($\text{RAG}^\text{2}$ and LongRAG), reflecting the improvements from decision optimization.

(2) Compared to adaptive RAG methods (SEAKR, DRAGIN, and Adaptive-RAG), our approach shows competitive time consumption, only slightly higher than Adaptive-RAG but lower than SEAKR and DRAGIN. Unlike SEAKR and DRAGIN, which require access to LLMs' output probability distributions, our method maintains adaptability for closed-source LLMs and API-based deployments. While both Adaptive-RAG and our method employ classifiers for decision optimization, our BERT-Base classifier (110M parameters) is more lightweight than Adaptive-RAG's T5-Large (770M parameters).

(3) Fig.~\ref{time_and_performance}.b demonstrates that our method achieves superior diagnostic performance. Overall, the proposed approach better balances efficiency and performance compared to baseline methods.

\section{Conclusion}
In this paper, we propose ICA-RAG, an adaptive retrieval decision optimization method for disease diagnosis that addresses the rigid retrieval strategy issue in traditional retrieval-augmented methods. ICA-RAG establishes a decision mechanism based on input information completeness to flexibly determine retrieval necessity, and introduces a retrieval and reranking strategy using document segmentation and mapping.
Experimental results demonstrate ICA-RAG's strong adaptability in complex clinical scenarios. Future work may explore further optimization of the retrieval process and ICA-RAG's application to other medical tasks.

\section*{Limitations}
Although our classification data annotation strategy is straightforward and effective, it still exhibits certain shortcomings in practical application. Due to the potential presence of repetitive content within the input patient information, LLMs may still arrive at a correct diagnosis even after masking a critical sentence. This can result in inaccurate annotation labels, necessitating manual inspection and revision on top of our proposed annotation strategy. Moreover, clinical medical texts, particularly EMRs, often contain abbreviations, synonyms, and aliases. And the manner in which identical patient information is recorded can vary significantly among different physicians, leading to a high degree of inconsistency. This issue to some extent hampers the search accuracy of our retrieval system. In the future, we aim to explore more effective preprocessing strategies for medical texts to enhance retrieval quality.

\section*{Ethical Consideration}
In this paper, we focus on the medical domain, specifically on enhancing the reliability and efficiency of retrieval-augmented generation (RAG) systems for disease diagnosis using large language models (LLMs). Our goal is to support better-informed decision-making by adaptively determining the necessity of information retrieval based on the information completeness of the input data. While our results demonstrate significant improvements in diagnostic accuracy and efficiency with the ICA-RAG framework, we need to stress that LLMs, even when augmented with retrieval mechanisms, should not be solely relied upon without the oversight of a qualified medical expert. The involvement of a physician or an expert is essential to validate the model’s recommendations and ensure a safe and effective decision-making process.   

Moreover, we acknowledge the profound ethical implications of deploying AI in healthcare. It is crucial to recognize that LLMs are not infallible and can produce erroneous outputs, even with advanced retrieval mechanisms. Transparency in how these models, including the decision-making process of ICA-RAG (e.g., why retrieval was or was not triggered), reach their conclusions, and incorporating continuous feedback from healthcare professionals are vital steps in maintaining the integrity and safety of medical practice. 

\bibliographystyle{IEEEtran}
\bibliography{ref}

\end{document}